\def\year{2017}\relax
\documentclass[letterpaper]{article} 
\usepackage{aaai17}  
\usepackage{times}  
\usepackage{helvet}  
\usepackage{courier}  
\usepackage{url}  
\usepackage{graphicx}  
\pdfoutput=1
\usepackage[cmex10]{amsmath}
\usepackage{amssymb}
\usepackage{amsthm}
\usepackage[linesnumbered,ruled]{algorithm2e}
\usepackage{subfig}
\usepackage{colortbl,booktabs}
\usepackage{threeparttable}

\DeclareMathOperator*{\argmin}{argmin}


\begin{document}
%
\title{Robust Kernelized Multi-View Self-Representations for Clustering by Tensor Multi-Rank Minimization}

\author{Yanyun Qu, Jinyan Liu\\
Fujian Key Laboratory of Sensing and Computing \\for Smart City, School of Information
Science \\and Engineering, Xiamen University, China\\ yyqu@xmu.edu.cn, jyliu2014@foxmail.com \\
\And Yuan Xie, Wensheng Zhang \\ State Key Laboratory of Intelligent Control\\ and Management of Complex Systems
\\Institute of Automation, Chinese Academy of Sciences \\ yuan.xie@ia.ac.cn, wensheng.zhang@ia.ac.cn
}

\maketitle
\begin{abstract}

Most recently, tensor-SVD is implemented on multi-view self-representation clustering and has achieved the promising results in many real-world applications such as face clustering, scene clustering and generic object clustering. However, tensor-SVD based multi-view self-representation clustering is proposed originally to solve the clustering problem in the multiple linear subspaces, leading to unsatisfactory results when dealing with the case of non-linear subspaces. To handle data clustering from the non-linear subspaces, a kernelization method is designed by mapping the data from the original input space to a new feature space in which the transformed data can be clustered by a multiple linear clustering method. In this paper, we make an optimization model for the kernelized multi-view self-representation clustering problem. We also develop a new efficient algorithm based on the alternation direction method and infer a closed-form solution. Since all the subproblems can be solved exactly, the proposed optimization algorithm is guaranteed to obtain the optimal solution. In particular, the original tensor-based multi-view self-representation clustering problem is a special case of our approach and can be solved by our algorithm. Experimental results on several popular real-world clustering datasets demonstrate that our approach achieves the state-of-the-art performance.

\end{abstract}

\section{Introduction}\label{intro}		
With the advent of big data, clustering attracts more and more attentions. In the past decades, great progresses have been made in clustering methods. The spectral clustering based methods are the mainstream clustering methods. There are two critical factors in spectral clustering, one is the subspace representation, and the other is the affinity matrix construction. The early spectral clustering methods focus on how to construct the affinity matrix \cite{spectral-clustering}. Recently, more and more concerns are given to subspace representation for clustering. Sparse subspace clustering \cite{SSC} tries to find a sparse representation for an instance, and the affinity matrix is constructed in the sparse subspace. The low-rank representation (LRR) method \cite{LRR} is a typical subspace clustering method, and it pays attention to a self-representation subspace, and the affinity matrix is constructed based on the self-representation coefficients.

Spectral clustering has been extended to the multi-view domain for machine learning tasks. In fact, many scientific data have heterogeneous features, which are collected from different domains or generated from different feature extractors. Here, we can give two typical application scenarios: 1) web-pages can be represented by using both page-text and hyperlinks pointing to them; 2) images can be described by different kinds of features, such as color, edge and texture. Each type of feature is regarded as a particular view, and multi-view features are combined for clustering. It has been proved that multi-view features could improve the clustering performance \cite{Survey,CCA}.

Extensive studies are made in multi-view clustering methods. The existing multi-view clustering methods are divided into three categories \cite{Survey}: 1) graph-based approaches, 2) co-training or co-regularized approaches, 3) subspace learning approaches. We will focus on the last category in this paper. Most existing multi-view subspace clustering (MSC) methods grasp the clustering cues only in a 2-order way and neglect the high order information, so that they may not make the best of the consensus property and the complementary property. Xie \textsl{et al.} \cite{t-SVD-MSC} made the multi-view features form a tensor, and utilized the self-representation strategy as a constraint and enforced the tensor multi-rank minimization on the objective function for clustering. It is referred to as t-SVD-MSC. This method has achieved the state-of-the-art multi-view clustering results on several popular image datasets.

However, it is originally proposed to handle the data from multiple linear subspaces on the input space, and it may not be suitable for the data from the non-linear subspace. Inspired by the work \cite{t-SVD-MSC,RKLRR}, we could use kernel-induced mapping to transform the data from the original input space to a feature space, in which the data reside in the union of multiple linear subspaces. Therefore, a new kernelized version of tensor-SVD multi-view self-representation model is proposed for clustering. We rewrite the optimization problem of tensor multi-view self-representation for clustering in the kernelized version, where the data matrix $X$ appears only in the form of inner product. In order to solve this optimization problem, which is actual a convex but nonsmooth objective function, we adopt the alternation direction method (ADM) \cite{LADM,ADM2,ADM3} with multiple blocks of variables. ADM can theoretically guarantee to achieve the global optimum.

The main contributions are three folds. 1) We firstly present a kernel version of multi-view self-representation subspace clustering, namely Kt-SVD-MSC, to deal with the case of non-linear subspaces. An optimization model is made for the kernel tensor based multi-view self-representation, and MSC is solved in the kernel induced feature subspace. 2) We develop a new optimization algorithm based on ADM to solve the optimization problem of Kt-SVD-MSC. We infer the analytical solutions for the subproblems with respect to $l_{2,1}$ norm and obtain the closed-form solutions. More interestingly, the original t-SVD-MSC becomes the special case of our method. Thus, the proposed algorithm can be used to solve the optimization problem of t-SVD-MSC. In addition, the kernel trick makes the computation efficient. 3) The proposed method achieves breakthrough clustering results on Scene-15 which is a challenging clustering dataset and achieves the state-of-the-art results on the other testing datasets, such as face clustering and generic object clustering, which demonstrates the effectiveness of our method.

The rest of this paper is organized as follows. The t-SVD-MSC is briefly introduced in section \ref{sec:t-svd-msc}. The proposed Kt-SVD-MSC is detailed in section \ref{sec:KLRC-MSC}. And experimental results are shown in section \ref{sec:experiment}. Conclusions are given in Section \ref{sec:conclusion}.

\section{Brief review of Tensor-SVD based multi-view subspace clustering}\label{sec:t-svd-msc}
In this section, we briefly review the general procedure of self-representation based subspace clustering methods. Low rank representation method (LRR) \cite{LRR} is a typical self-representation based subspace clustering method, and many self-representation methods are the variant of LRR \cite{LRR}.
Suppose $\mathbf{X}=[\mathbf{x}_1, \mathbf{x}_2, \ldots, \mathbf{x}_N] \in \mathbb{R}^{{d}\times N} $ is the data matrix, whose column is a sample vector of ${d}$ dimensions in the feature space. LRR can be formulated as the following optimization problem,
\begin{equation}\label{LRR_problem}
    \begin{aligned}
    &\min_{\mathbf{Z}, \mathbf{E}} \lambda ||\mathbf{E}||_{2,1} + || \mathbf{Z}||_{\ast},\\
    &\text{s.t.} \quad \mathbf{X} = \mathbf{X}\mathbf{Z} + \mathbf{E}\\
    \end{aligned}
\end{equation}
where $\mathbf{Z} = [\mathbf{z}_1, \mathbf{z}_2, \ldots, \mathbf{z}_N] \in \mathbb{R}^{N\times N}$ is the coefficient matrix with each $\mathbf{z}_i$ being the new coding of sample $\mathbf{x}_i$. $\|\quad \|_{\ast}$ is the nuclear norm, $\|\quad \|_{2,1}$ denotes the $\textit{l}_{2,1}$ norm of a matrix. $\mathbf{E}$ is the corresponding noise matrix. After obtaining the self-representation matrix $\mathbf{Z}$, the affinity matrix is constructed as,
\begin{equation}\label{A_problem}
    \mathbf{A} = \frac{1}{2}( |\mathbf{Z}| + |\mathbf{Z}^{\mathrm{T}}|)
\end{equation}
where $|\quad|$ represents the absolute operator. And then, spectral clustering is utilized with the affinity matrix $\mathbf{A}$ to generate the clustering result.

LRR can be extended to multi-view clustering. Let $\mathbf{X}^{(v)}$ denote the feature matrix of the ${v}$-th view and $\mathbf{Z}^{(v)}$ denote the ${v}$-th learned subspace representation. The multi-view subspace clustering problem can be reformulated as,
\begin{equation}\label{multiview_LRR_problem}
    \begin{aligned}
    &\min_{\mathbf{Z}^{(v)}, \mathbf{E}^{(v)}} \sum_{v=1}^{V} (\lambda ||\mathbf{E}^{(v)}||_{2,1} + ||\mathbf{Z}^{(v)}||_{\ast}),\\
    &\text{s.t.} \quad \mathbf{X}^{(v)} = \mathbf{X}^{(v)}\mathbf{Z}^{(v)} + \mathbf{E}^{(v)}, v=1,\ldots, V,\\
    \end{aligned}
\end{equation}
where ${V}$ denotes the number of all views and $\mathbf{E}^{(v)}$ is the noise matrix corresponding to the ${v}$-th view. From (\ref{multiview_LRR_problem}), the multi-view clustering can use LRR in a direct and simple way. After obtaining the multi-view subspace representation, the affinity matrix can be constructed as $\mathbf{A} = \frac{1}{V}\sum_{v=1}^{V}( |\mathbf{Z}^{(v)}| + |\mathbf{Z}^{(v)^{\mathrm{T}}}|)/2$.

The limitation of (\ref{multiview_LRR_problem}) is that it treats each subspace representation independently, neglecting the complementarity of different views. Thus, a tensor based clustering method is proposed in which a tensor nuclear norm (TNN) is used as a regularization term instead of the matrix nuclear norm.
\begin{figure}[htb]
\centering
\includegraphics[width=0.4\textwidth]{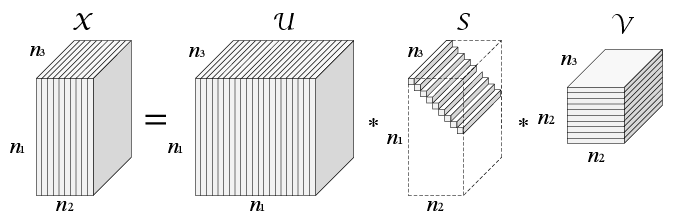}
\caption{The t-SVD of an $n_{1} \times n_{2} \times n_{3}$ tensor.}
\label{fig:tSVD}
\end{figure}
A tensor nuclear norm is defined as,
\begin{equation}
\label{fml:gtnn}
||\boldsymbol{\mathcal{X}}||_{\circledast} :=\sum_{i=1}^{\mathrm{min}(n_{1},n_{2})}\sum_{k=1}^{n_{3}}|{ \boldsymbol{\mathcal{S}}}_{f}(i,i,k)|,
\end{equation}
where $\boldsymbol{\mathcal{X}}$ is a tensor. The tensor singular value decomposition is $\boldsymbol{{\mathcal{X}}}=\boldsymbol{{\mathcal{U}}}*\boldsymbol{{\mathcal{S}}}*\boldsymbol{{\mathcal{V}}}^{\mathrm{T}}$, where $\boldsymbol{{\mathcal{U}}}$ and $\boldsymbol{{\mathcal{V}}}$ are orthogonal tensors of size $n_{1} \times n_{1} \times n_{3}$ and $n_{2} \times n_{2} \times n_{3}$ respectively, and $\boldsymbol{{\mathcal{S}}}$ is an f-diagonal tensor of size $n_{1} \times n_{2} \times n_{3}$, as shown in Fig.\ref{fig:tSVD} .
\begin{figure}[htb]
\centering
\includegraphics[width=0.4\textwidth]{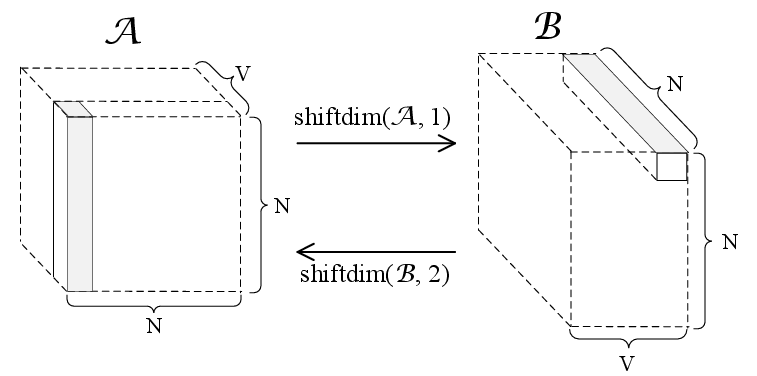}
\caption{The rotated coefficient tensor in our approach.}
\label{fig:shift-tSVD}
\end{figure}
And the tensor multi-rank minimization problem for multi-view self-representation clustering is formulated as,
\begin{equation}\label{tsvd_msc_problem}
    \begin{aligned}
    &\min_{\mathbf{Z}^{(v)}, \mathbf{E}^{(v)}} \lambda ||\mathbf{E}||_{2,1} + ||\boldsymbol{ \mathcal{Z}}||_{\circledast},\\
    &\text{s.t.} \quad \mathbf{X}^{(v)} = \mathbf{X}^{(v)}\mathbf{Z}^{(v)} + \mathbf{E}^{(v)}, v=1,\ldots, V,\\
    &\qquad \boldsymbol{\mathcal{Z}} = \Phi(\mathbf{Z}^{(1)}, \mathbf{Z}^{(2)}, \ldots, \mathbf{Z}^{(V)}),\\
    &\qquad \mathbf{E} = [\mathbf{E}^{(1)}; \mathbf{E}^{(2)}; \ldots, \mathbf{E}^{(V)}],
    \end{aligned}
\end{equation}
where the function $\Phi(\cdot)$ constructs the tensor $\boldsymbol{\mathcal{Z}}$ by stacking different representation $\mathbf{Z}^{(v)}$ to a $3$-mode tensor. As illustrated in Fig. \ref{fig:shift-tSVD}, we rotates its dimensionality to $N\times V \times N$. In Fig. \ref{fig:shift-tSVD}, the self-representation coefficient(mode-$1$ fiber marked by the gray bar in Tensor $\mathcal{A}$) is transformed into the mode-$3$ fiber marked by the gray bar in Tensor $\mathcal{B}$. The notation of tensor is detailed in \cite{t-SVD-MSC}.

\section{The Proposed Method}\label{sec:KLRC-MSC}
\begin{figure*}[!htb]
\centering
\includegraphics[width=0.8\textwidth]{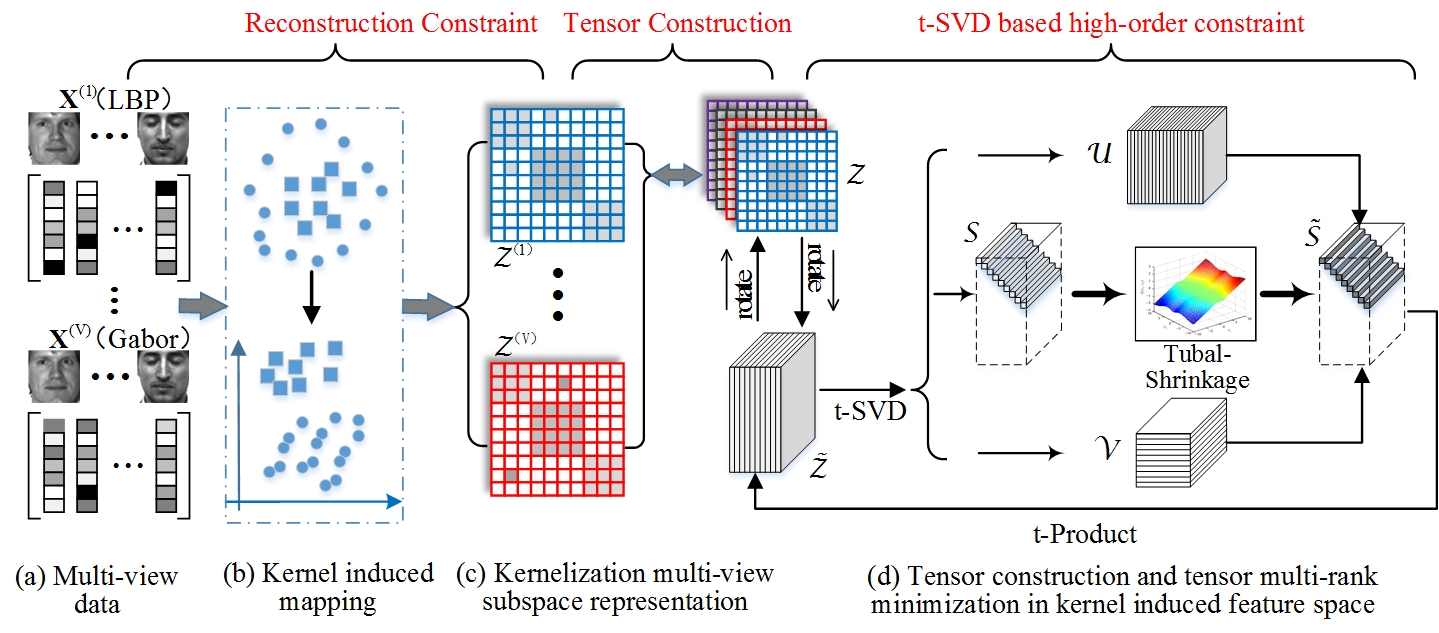}
\caption{The framework of our method. a) Feature extraction for multi-view representation, obtaining the multi-view features $\{\mathbf{X}^{(v)}\}_{v=1}^{V}$ , b) Feature transformation from the input space to the new feature space,  c) Multi-view self-representation in the kernel induced feature space, d) Tensor construction by stacking the kernel multi-view self-representation $\{\mathbf{Z}^{(v)}\}_{v=1}^{V} $, and then solving the subspace representations optimal solution by using t-SVD based tensor multi-rank minimization }
\label{fig:framework}
\end{figure*}

In this section, we give the problem statement of kernelized multi-view self-representation by tensor multi-rank minimization.
The framework of the proposed method is illustrated in Fig.\ref{fig:framework}.

Let $\mathbf{X}^{(v)}$ denote the $v$-th view feature matrix whose column is the feature vector of a sample in the original subspace. We transform a feature vector in the original subspace to the non-linear feature subspace by using a kernel induced latent mapping function. In the feature subspace, t-SVD-MSC is implemented on the new feature matrix. In other words, self-representation is implemented on the new feature vectors in the $v$-th view feature subspace. And then the $V$-view subspace representation matrices are obtained, denoted by $\mathbf{Z}^{(1)},\cdots,\mathbf{Z}^{(V)}$. They are merged to constructed a $3$-order tensor $\boldsymbol{\mathcal{Z}}$. According to t-SVD-MSC \cite{t-SVD-MSC}, this tensor needs to be rotated so as to keep self-representation coefficient in Fourier domain. After that, the rotated tensor $\tilde{\boldsymbol{\mathcal{Z}}}$ is efficiently updated by tensor-SVD based tensor nuclear norm minimization in order to grasp the high order information of the multi-view representation. Sequently, each $\mathbf{Z}^{(v)}$ will be updated by using self-representation method. These operations are executed iteratively until convergence is achieved. The challenging task is how to solve the solution $\mathbf{Z}^{(v)}$  under the kernel induced feature subspace. In this paper, we utilize the kernel strategy to avoid finding the explicit non-linear mapping function and high complication complexity in a high-dimension subspace.

Following the existing kernel-based method \cite{kernelized-method1,kernelized-method2}, for the $v$-th view, let $\mathbf{K}^{(v)} \in \mathbb{R}^{N\times N}$ denote the kernel matrix, and let $ker^{(v)}(\mathbf{x}, \mathbf{y})$ denote the kernel function, we have:
\begin{equation}\label{kernel-function}
    \mathbf{K}^{(v)}_{ij} = ker^{(v)}(\mathbf{x}^{(v)}_i, \mathbf{x}^{(v)}_j), \quad \forall i, j = 1, \ldots, N.
\end{equation}
where the kernel function $ker^{(v)}(\mathbf{x}, \mathbf{y})$ induces a mapping $\psi^{(v)}: \mathbb{R}^{d^{(v)}}\rightarrow \mathcal{F}^{(v)}$ (usually implicitly defined), in which $\mathcal{F}^{(v)}$ is referred to as the new feature space \cite{kernel-space}. Given the mapping $\psi^{(v)}$, function $ker^{(v)}(\cdot,\cdot)$ can be rewritten as:
\begin{equation}\label{kernel-function1}
    ker^{(v)}(\mathbf{x}, \mathbf{y}) = \psi^{(v)}(\mathbf{x})^{T}\psi^{(v)}(\mathbf{y}).
\end{equation}
Let's define $\Psi^{(v)}(\mathbf{X}^{(v)}) = [\psi^{(v)}(\mathbf{x}^{(v)}_1), \ldots, \psi^{(v)}(\mathbf{x}^{(v)}_N)]$, and the kernel matrix of the $v$-th view can be calculated as:
\begin{equation}\label{kernel-view}
    \mathbf{K}^{(v)} = \Psi^{(v)}(\mathbf{X}^{(v)})^{T}\Psi^{(v)}(\mathbf{X}^{(v)}).
\end{equation}

Furthermore, the reconstruction error in feature space $\mathcal{F}^{(v)}$ can be expressed as:
\begin{equation}\label{recon-constraint-kernel}
    \mathbf{E}^{(v)} = \Psi^{(v)}(\mathbf{X}^{(v)}) - \Psi^{(v)}(\mathbf{X}^{(v)}) \mathbf{Z}^{(v)}.
\end{equation}
Thus, (\ref{tsvd_msc_problem}) can be converted to the non-linear version of t-SVD-MSC in the feature subspace $\mathcal{F}^{(v)}$ as:
\begin{equation}\label{ktsvd_msc_problem1}
    \begin{aligned}
\min_{\mathbf{Z}^{(v)}} \quad &\lambda \sum_{v=1}^{V}||\Psi^{(v)}(\mathbf{X}^{(v)}) - \Psi^{(v)}(\mathbf{X}^{(v)}) \mathbf{Z}^{(v)}||_{2,1} + ||\boldsymbol{ \mathcal{Z}}||_{\circledast},\\
    &\boldsymbol{\mathcal{Z}} = \Phi(\mathbf{Z}^{(1)}, \mathbf{Z}^{(2)}, \ldots, \mathbf{Z}^{(V)}).
    \end{aligned}
\end{equation}
The error term can be rewritten as:
\begin{equation}
\label{rewritten-error-term}
    \begin{array}{ccl}
       & &\|\Psi^{(v)}(\mathbf{X}^{(v)}) - \Psi^{(v)}(\mathbf{X}^{(v)}) \mathbf{Z}^{(v)}\|_{2,1}\\
        & = &\|\Psi^{(v)}(\mathbf{X}^{(v)}) \mathbf{P}^{(v)}\|_{2,1}\\
       & = &\sum_{i=1}^{N} (\mathbf{p}_i^{(v)^{T}}  \mathbf{K}^{(v)} \mathbf{p}^{(v)}_i)^{\frac{1}{2}},
    \end{array}
\end{equation}
where $\mathbf{P}^{(v)} = \mathbf{I} - \mathbf{Z}^{(v)} \in \mathbb{R}^{N \times N}$ and $\mathbf{P}^{(v)} = [\mathbf{p}^{(v)}_1, \ldots, \mathbf{p}^{(v)}_N]$. Let's define $ h^{(v)}(\mathbf{P}^{(v)}) = \sum_{i=1}^{N} \sqrt{\mathbf{p}_i^{(v)^{T}}  \mathbf{K}^{(v)} \mathbf{p}^{(v)}_i)}$. Consequently, the problem (\ref{ktsvd_msc_problem1}) can be reformulated as:
\begin{equation}\label{Kt-SVD-MSC-real-problem1}
    \begin{aligned}
    \min_{\mathbf{Z}^{(v)}, \mathbf{P}^{(v)}} \quad &\lambda \sum_{v=1}^{V} h^{(v)}(\mathbf{P}^{(v)}) + ||\boldsymbol{ \mathcal{Z}}||_{\circledast},\\
    &\boldsymbol{\mathcal{Z}} = \Phi(\mathbf{Z}^{(1)}, \mathbf{Z}^{(2)}, \ldots, \mathbf{Z}^{(V)}),\\
    &\mathbf{P}^{(v)} = \mathbf{I} - \mathbf{Z}^{(v)}, v=1,\ldots,V.
    \end{aligned}
\end{equation}
Obviously, after converting the non-linear version (\ref{ktsvd_msc_problem1}) into the formulation in (\ref{Kt-SVD-MSC-real-problem1}), we can solve this optimization problem by using the kernel trick without computing a explicit non-linear mapping function $\Psi^{(v)}(\mathbf{X}^{(v)})$ and high computational complexity.

\subsection{Optimization}

In this subsection, we solve the problem (\ref{Kt-SVD-MSC-real-problem1}) by using alternating direction minimizing(ADM)\cite{LADM}. According to the Augmented Lagrange Multiplier (ALM) \cite{alm-lin}, we introduce multiple blocks of Lagrange multipliers: the matrix $\mathbf{Y}_{v}, (v=1,2,\cdots, V)$ and the tensor $\boldsymbol{\mathcal{W}}$ , and introduce the auxiliary tensor variable $\boldsymbol{\mathcal{G}}$ to replace $\boldsymbol{ \mathcal{Z}}$ in the TNN regularization term \cite{t-SVD-MSC}. The optimization problem can be transferred to minimize the following unconstrained problem:
\begin{equation}\label{t-SVD-MSC}
    \begin{aligned}
    &\boldsymbol{\mathcal{L}}(\mathbf{Z}^{(1)}, \ldots, \mathbf{Z}^{(V)}; \mathbf{P}^{(1)}, \ldots, \mathbf{P}^{(V)};\mathbf{Y}_{1}, \ldots, \mathbf{Y}_{V};\boldsymbol{\mathcal{W}};\boldsymbol{\mathcal{G}} ) \\
    &= \|\boldsymbol{ \mathcal{G}}\|_{\circledast}+ \lambda \sum_{v=1}^{V} h^{(v)}(\mathbf{P}^{(v)}) \\
    &+ \sum_{v=1}^{V} \bigg(\langle \mathbf{Y}_{v}, \mathbf{I}^{(v)} - \mathbf{Z}^{(v)} - \mathbf{P}^{(v)} \rangle + \frac{\mu}{2} \|\mathbf{I}^{(v)} - \mathbf{Z}^{(v)} - \mathbf{P}^{(v)}\|_{F}^{2} \bigg)\\
    & + \langle \boldsymbol{\mathcal{W}}, \boldsymbol{\mathcal{Z} - \boldsymbol{\mathcal{G}}} \rangle + \frac{\rho}{2} ||\boldsymbol{\mathcal{Z} - \boldsymbol{\mathcal{G}}}||_{F}^{2}.
    \end{aligned}
\end{equation}
where $\mu $ and $\rho$ are actually the penalty parameters, which are adjusted by using adaptive updating strategy as suggested in \cite{LADM}.

ADM is adopted for updating $\mathbf{Z}^{(v)}$, $\mathbf{P}^{(v)}$, $\boldsymbol{\mathcal{G}}$ and $\mathbf{Y}_{v}$ $(v=1,2,\cdots, V)$. The detailed procedure for the $v$-th view can be partitioned into four steps alternatingly.
\paragraph{1. $\mathbf{Z}^{(v)}$-subproblem:} When $\mathbf{P}$, and $\boldsymbol{\mathcal{G}}$ are fixed, we define $\Phi_{(v)}^{-1}(\boldsymbol{\mathcal{W}}) = \mathbf{W}^{(v)}$ and $\Phi_{(v)}^{-1}(\boldsymbol{\mathcal{G}}) = \mathbf{G}^{(v)}$, and then we will solve the following subproblem for updating the subspace representation $\mathbf{Z}^{(v)}$:
\begin{equation}\label{z-subproblem}
    \begin{aligned}
    \min_{\mathbf{Z}^{(v)}} &\quad \langle \mathbf{Y}_{v}, \mathbf{I} - \mathbf{Z}^{(v)} - \mathbf{P}^{(v)} \rangle + \frac{\mu}{2} ||\mathbf{I} - \mathbf{Z}^{(v)} - \mathbf{P}^{(v)}||_{F}^{2}\\
    & + \langle \mathbf{W}^{(v)}, \mathbf{Z}^{(v)} - \mathbf{G}^{(v)} \rangle + \frac{\rho}{2} ||\mathbf{Z}^{(v)} - \mathbf{G}^{(v)}||_{F}^{2}.
    \end{aligned}
\end{equation}
By setting the derivative of (\ref{z-subproblem}) to zero, the closed-form of $\mathbf{Z}^{(v)}$ can be obtained by
\begin{equation}\label{z-subproblem-solution}
    \begin{aligned}
    \mathbf{Z}^{(v)^{\ast}} = & (\mu\mathbf{I} +\mathbf{Y}_{v}+\rho\mathbf{G}^{(v)}-\mu\mathbf{P}^{(v)}-\mathbf{W}^{(v)})/(\mu+\rho).
    \end{aligned}
\end{equation}
The solution is denoted by $\mathbf{Z}_{t+1}^{(v)}$.
\paragraph{2. $\mathbf{P}^{(v)}$-subproblem:} After obtaining $\mathbf{Z}_{t+1}^{(v)}$, we solve the optimization solution $\mathbf{P}^{(v)}$ by fixing the other parameters.
\begin{equation}\label{P-subproblem}
\begin{aligned}
 &  \min_{\mathbf{P}^{(v)}}\quad \lambda h^{(v)}(\mathbf{P}^{(v)}) + \langle \mathbf{Y}_{v}, \mathbf{I} - \mathbf{Z}^{(v)} - \mathbf{P}^{(v)} \rangle
   + \frac{\mu}{2} \|\mathbf{I} - \mathbf{Z}^{(v)} - \mathbf{P}^{(v)}\|_{F}^{2}\\
&= \min_{\quad \mathbf{P}^{(v)}} \lambda h^{(v)}(\mathbf{P}^{(v)}) +\frac{\mu}{2} ||\mathbf{I} - \mathbf{Z}^{(v)} - \mathbf{P}^{(v)}+\mathbf{Y}_{v}/\mu||_{F}^{2}\\
&= \min_{\quad \mathbf{P}^{(v)}} \lambda h^{(v)}(\mathbf{P}^{(v)}) +\frac{\mu}{2} ||\mathbf{P}^{(v)}-\mathbf{C}^{(v)}||_{F}^{2}
\end{aligned}
\end{equation}
where $\mathbf{C}^{(v)}=\mathbf{I} -\mathbf{Z}^{(v)}+\mathbf{Y}_{v}/\mu $, and we have dropped the terms that are irrelevant to $\mathbf{P}^{(v)}$.

Note that it is nontrivial to solve the problem in (\ref{P-subproblem}), where the objective function is convex but nonsmooth.
In order to explain how to solve the second subproblem, we formulate the SVD of the $v$-th view kernel matrix as,
\begin{equation}
    \begin{aligned}
    \mathbf{K}^{(v)}=\mathbf{V}^{(v)}\Sigma^{(v)2}\mathbf{V}^{(v)^{T}}
    \end{aligned}
\end{equation}
where $\Sigma^{(v)}=\mathbf{diag}(\sigma_1^{(v)},\dots,\sigma_\mathbf{r_K}^{(v)},0,\dots,0)$ and $r_{K}$ is the rank of Matrix $\mathbf{K}^{(v)}$.
And we define $\mathbf{c}_{i}^{(v)}$ (respectively, $\mathbf{p}_{i}^{(v)}$) as the $i$th column of $\mathbf{C}^{(v)}$ (respectively, $\mathbf{P}^{(v)}$) and scalar $\tau=\mu/\lambda$. We have the solution $\mathbf{p}^{(v)}_{i}$  according to \cite{RKLRR}:
\begin{equation}\label{P-subproblem-solution}
    \begin{aligned}
    \mathbf{p}_{i}^{(v)^{\ast}} =
    \begin{cases}
\hat{\mathbf{p}}^{(v)},if\quad||[1/\sigma_1^{(v)},\dots,1/\sigma_\mathbf{r_K}^{(v)}]\circ\mathbf{t}_{u}^{(v)}||>1/\tau \\
\mathbf{c}_{i}^{(v)}-\mathbf{V_K}^{(v)}\mathbf{t}_{u}^{(v)},\mathbf{otherwise}
    \end{cases}
    \end{aligned}
\end{equation}
where $\circ$ is the element multiplication operator, and $\mathbf{a}\circ\mathbf{b}$ is a new vector whose $i$th element is $\mathbf{a}_i\mathbf{b}_i$.

Let's define $\mathbf{t}_{u}^{(v)}=\mathbf{V_K}^{(v)^{T}}\mathbf{c}_{i}^{(v)} \in \mathbb{R}^{{r}_{K}^{(v)}} $, where $\mathbf{V_K}^{(v)} \in \mathbb{R}^{N \times {r}_{K}^{(v)}}$ is formed by the first ${{r}_{K}^{(v)}}$ columns of $\mathbf{V}^{(v)}$, and the vector $\hat {\mathbf{p}}^{(v)}$ is defined as
\begin{equation}
    \begin{aligned}
    \hat{\mathbf{p}}^{(v)}=\mathbf{c}_{i}^{(v)}-\mathbf{V_K}^{(v)}([\frac{\sigma_1^{(v)2}}{\alpha^{(v)}+\sigma_1^{(v)2}},\dots,\frac{\sigma_\mathbf{r_K}^{(v)2}}{\alpha^{(v)}+\sigma_\mathbf{r_K}^{(v)2}}]^{T}\circ\mathbf{t}_{u}^{(v)})
    \end{aligned}
\end{equation}
where $\alpha^{(v)}$ is a positive scalar, satisfying
\begin{equation}\label{alpha-subproblem}
    \begin{aligned}
\mathbf{t}_{u}^{{(v)}^{T}}\mathbf{diag}(\{\frac{\sigma_\mathbf{i}^{(v)2}}{(\alpha^{(v)}+\sigma_\mathbf{r_i}^{(v)2})^2}\}_{1\le\mathbf{i}\le{r}_{K}^{(v)} })\mathbf{t}_{u}^{(v)}=\frac{1}{\tau^2}
    \end{aligned}
\end{equation}
In particular, when $\|[1/\sigma_1^{(v)},\dots,1/\sigma_\mathbf{r_K}^{(v)}]\circ\mathbf{t}_{u}^{(v)}\|>1/\tau$, the equation in (\ref{alpha-subproblem}) (with respect to $\alpha$) has a unique positive root, which can be obtained by the bisection method \cite{Bisection}.

\paragraph{3. $\mathbf{G}^{(v)}$-subproblem:}
When $\mathbf{Z}^{(v)},(v=1,2,\dots,\mathbf{V})$, are fixed, we solve the following subproblem in order to update the tensor $\mathcal{G}$,
\begin{equation}\label{G-subproblem}
    \boldsymbol{\mathcal{G}}^{\ast} = \argmin_{\boldsymbol{\mathcal{G}}} || \boldsymbol{\mathcal{G}} ||_{\circledast} + \frac{\rho}{2} ||\boldsymbol{\mathcal{G}} - (\boldsymbol{\mathcal{Z}} + \frac{1}{\rho}\boldsymbol{\mathcal{W}})||_{F}^{2}.
\end{equation}
Let's define $\boldsymbol{\mathcal{F}}=\boldsymbol{\mathcal{Z}} + \frac{1}{\rho}\boldsymbol{\mathcal{W}}$. According to \cite{t-SVD-MSC}, let $\mathrm{fft}(\boldsymbol{\mathcal{F}},[~], 3)$ denote the Fourier transform along the third dimension and $\mathrm{ifft}(\boldsymbol{\mathcal{G}}_{f},[~], 3)$ denote the inverse Fourier transform along the third dimension . The solution of the optimization problem (\ref{G-subproblem}) can be achieved through Algorithm \ref{TNN}.The detail can be found in \cite{t-SVD-MSC}.

\begin{algorithm}
\SetAlgoLined
\caption{t-SVD based Tensor Nuclear Norm Minimization}\label{TNN}
\KwIn{Observed tensor $\boldsymbol{\mathcal{F}}\in \mathbb{R}^{n_1 \times n_2 \times n_3}$}
\KwOut{tensor $\boldsymbol{\mathcal{G}}$}
\BlankLine
$\boldsymbol{\mathcal{F}}_{f} = \mathrm{fft}(\boldsymbol{\mathcal{F}},[~], 3)$\;
\For{$j=1:n_3$}
{
    $[\boldsymbol{\mathcal{U}}_{f}^{(j)}, \boldsymbol{\mathcal{S}}_{f}^{(j)}, \boldsymbol{\mathcal{V}}_{f}^{(j)}] = \mathrm{SVD}(\boldsymbol{\mathcal{F}}_{f}^{(j)})$\;

    $\boldsymbol{\mathcal{J}}_{f}^{(j)} = \mathrm{diag}\{(1 - \frac{\tau'}{\boldsymbol{\mathcal{S}}_{f}^{(j)}(i,i)})_{+}\}, i=1, \ldots, \min(n_1, n_2)$\;

    $\boldsymbol{\mathcal{S}}_{f,n_3}^{(j)} = \boldsymbol{\mathcal{S}}_{f}^{(j)} \boldsymbol{\mathcal{J}}_{J}^{(j)}$\;

    $\boldsymbol{\mathcal{G}}_{f}^{(j)} = \boldsymbol{\mathcal{U}}_{f}^{(j)} \boldsymbol{\mathcal{S}}_{f,n_3}^{(j)} \boldsymbol{\mathcal{V}}_{f}^{(j)^{\mathrm{T}}}$\;
}
$\boldsymbol{\mathcal{G}} = \mathrm{ifft}(\boldsymbol{\mathcal{G}}_{f},[~], 3)$\;
\textbf{Return} tensor $\boldsymbol{\mathcal{G}}$.
\end{algorithm}
\paragraph{4. Updating the Lagrange multipliers $\mathbf{Y}^{(v)}$ , $\boldsymbol{\mathcal{W}}$ , $\mu$, and $\rho$ :}
\begin{equation}\label{Y-subproblem}
    \mathbf{Y}_{v}^{\ast}  = \mathbf{Y}_{v} + \mu(\mathbf{I}-\mathbf{Z}^{(v)} - \mathbf{P}^{(v)}), \\
\end{equation}
\begin{equation}\label{W-subproblem}
    \boldsymbol{\mathcal{W}}^{\ast}  = \boldsymbol{\mathcal{W}} + \rho(\boldsymbol{\mathcal{Z}} - \boldsymbol{\mathcal{G}}), \\
\end{equation}
\begin{equation}\label{mu-subproblem}
    \mu  = \min(\eta \mu, \mu_{\max}), \\
\end{equation}
\begin{equation}\label{rho-subproblem}
    \rho  = \min(\eta \rho, \rho_{\max}).
\end{equation}
where $\mu_{\max}$ and $\rho_{\max}$ are presetting parameters.

We summarize the algorithm for Kt-SVD-MSC in Algorithm \ref{proposed-KMSC}.
\begin{tiny}
\begin{algorithm}[htb]
\SetAlgoLined
\caption{Kt-SVD-MSC}
\label{proposed-KMSC}
\KwIn{Multi-view feature matrices: $\mathbf{X}^{(1)}, \mathbf{X}^{(2)}, \ldots, \mathbf{X}^{(V)}$, $\lambda$, cluster number $C$, and the SVD of $\mathbf{K^{(v)}}$}
\KwOut{Clustering results $\mathcal{C}$}
\BlankLine
Initialize $\mathbf{Z}^{(v)} = \mathbf{Y}_{v} =  \mathbf{0}$, $i = 1,\ldots, V$;  $\mathbf{P}^{(v)}=I_n$ $\boldsymbol{\mathcal{G}} = \boldsymbol{\mathcal{W}} = \mathbf{0}$;\\
$\mu =\rho = 10^{-5}$, $\eta = 2$, $\mu_{\max} =10^{13}, \rho_{\max} = 10^{13}$, $\varepsilon = 10^{-7}$; \\
\While {not converge}
{
    \For {$v=1:V$ }
   {
    // {Solving $\mathbf{Z}$}\\
    Update $\{\mathbf{Z}^{(v)}\}_{v=1}^{V}$ by using (\ref{z-subproblem-solution});\\
   }
    // {Solving $\mathbf{P}$}\\
    Update $\mathbf{P}$ by solving (\ref{P-subproblem-solution});\\
   \For {$v=1:V$ }
   {
    // {Solving $\mathbf{Y}$}\\
    Update $\{\mathbf{Y}^{(v)}\}_{v=1}^{V}$ by using (\ref{Y-subproblem});\\
   }

    Obtain $\boldsymbol{\mathcal{Z}} = \Phi(\mathbf{Z}^{(1)}, \mathbf{Z}^{(2)}, \ldots, \mathbf{Z}^{(V)})$\;
    // {Solving $\mathbf{G}$};\\
    Update $\boldsymbol{\mathcal{G}}$ by using Algorithm \ref{TNN};  \\
    Update $\boldsymbol{\mathcal{W}}$ by using (\ref{W-subproblem});\\
    Update parameters $\mu$ and $\rho$ by using (\ref{mu-subproblem}) and (\ref{rho-subproblem}), respectively;\\
    Check the convergence conditions:\\
   \qquad $||\mathbf{I} - \mathbf{Z}^{(v)} - \mathbf{P}^{(v)}||_{\infty} < \varepsilon$ and \\
     \qquad $||\mathbf{Z}^{(v)} - \mathbf{G}^{(v)}||_{\infty} < \varepsilon$\;
}
Obtain the affinity matrix by\\
 \qquad $\mathbf{A} = \frac{1}{V}\sum_{v=1}^{V} (|\mathbf{Z}^{(v)}| + |\mathbf{Z}^{(v)^{\mathrm{T}}}|)$/2\;
Apply the spectral clustering method with the affinity matrix $\mathbf{A}$\;
\textbf{Return} Clustering result $\mathcal{C}$.
\end{algorithm}
\end{tiny}
\section{Experimental Results and Analysis}\label{sec:experiment}

In this section, in order to estimate the proposed method, we implement it on six challenging image clustering datasets: {\it Yale}\footnote{https://cvc.yale.edu/projects/yalefaces/yalefaces.html}, {\it Extended YaleB}\footnote{https://cvc.yale.edu/projects/yalefacesB/yalefacesB.html},  {\it ORL}\footnote{http://www.uk.research.att.com/facedatabase.html}, {\it Notting-Hill} \cite{notting-hill}, {\it Scene-15}\footnote{http://www-cvr.ai.uiuc.edu/ponce\_grp/data/}, {\it COIL-20}\footnote{http://www.cs.columbia.edu/CAVE/software/softlib/} . The first three datasets are three typical face datasets, and Nottting-Hill is a video face dataset. Scene-$15$ are for scene clustering, and COIL-$20$ is for generic object clustering. Table \ref{dataset} shows the details of the six datasets, such as the number of images and the number of classes. We use six popular criteria to evaluate a clustering method: Normalized Mutual Information (NMI), Accuracy (ACC), Adjusted Rank index (AR), F-score, Precision and Recall. For all the criteria, the higher the score is, the better the performance is. In all datasets, we use the average of ten runs as the final results. In the Section 1 of \textit{supplemental material}, we present more clustering results, such as the mean and standard deviation of the six criteria. All experiments are implemented in Matlab on a workstation with $4$GHz CPU, $32$GB RAM, and TITANX GPU($12$GB caches). To promotethe culture of reproducible research, source codes  accompanying this work can be achieved at http://pan.baidu.com/s/1bp4JUqz (password: m9qo).
\begin{table}[!ht]
\footnotesize
\begin{centering}
\begin{threeparttable}[]
\caption{Statistics of different test datasets}\label{dataset}
\tabcolsep=4pt
\begin{minipage}{12cm}
\begin{tabular}{@{}lccc}
\toprule[2pt] 
Dataset         &Images &Objective &Clusters  \\ \hline
Yale            & 165  & Face  & 15  \\
Extended YaleB     &640  &Face &10 \\
ORL                &400  &Face &10 \\
Notting-Hill        &4660 &Face &5 \\
Scene-15           &4485 &Scene &15 \\
COIL-20            &1440 &Generic Object &20 \\\hline \bottomrule[1pt]
\end{tabular}
\end{minipage}
\end{threeparttable}
\end{centering}
\end{table}

We take different strategies for clustering to deal with different applications. As for face clustering, three types of features are extracted: intensity, LBP \cite{LBP}, and Gabor \cite{Gabor}. The LBP feature is extracted from an image with the sampling size of $8$ pixels and the blocking number of $7\times 8$. The Gabor feature is extracted with only one scale $\lambda = 4$ at four orientations $\theta = \{0^{o}, 45^{o}, 90^{o}, 135^{o}\}$. Thus, the LBP feature is of $3304$ dimensions, and Gabor feature is of 6750 dimensions. As for Scene-$15$, three types of features are extracted: Pyramid histograms of visual words (PHOW)\cite{PHOW}, Pairwise rotation invariant co-occurrence local binary pattern feature (PRI-CoLBP)\cite{PRI-CoLBP}, and CENsus TRansform hISTogram (CENTRIST)\cite{CENTRIST}. PHOW is extracted with dense sampling step of $8$ pixels and $300$ words and is of $1800$ dimensions. PRI-CoLBP is of 1180 dimensions.  Three level features are extracted to form CENTRIST, which contains $1,5,25$ blocks respectively. CENTRIST is of $1240 $ dimensions. COIL-$20$ uses the same features as those used for face clustering. And the feature are extracted from a normalized image with size of $32\times 32$.

In the experiments, we implement a linear kernel on each view on {\it Yale} and {\it ORL} and implement linear kernels and Gaussian kernels on different views on {\it Extended YaleB, Scene-15, Notting-Hill and COIL-20}. We compare our approach with eight representative clustering methods: the standard spectral clustering algorithm with the most informative view (SPCbest), LRR algorithm with the most informative view (LRRbest), robust multi-view spectral clustering via low-rank and sparse decomposition (RMSC)\cite{graph-based4}, diversity-induced multi-view subspace clustering (DiMSC) \cite{DiMSC}, low-rank tensor constrained multi-view subspace clustering (LTMSC) \cite{LTMSC}, the tensor-SVD multi-view clustering method with tensor multi-rank minimization (t-SVD-MSC)\cite{t-SVD-MSC}, the tensor-SVD with unrotated coefficient tensor (Ut-SVD-MSC), and exclusivity-consistency regularized MSC (ECMSC) \cite{ECMSC}. The first two methods are the single view baselines. The RMSC, DiMSC, and LTMSC are three state-of-the-art methods in multi-view clustering. As for ECMSC, only  clustering results on {\it Yale, Extended YaleB, ORL} were published. Thus, we run the code provided by the authors on {\it Notting Hill, Scene-15 and COIL-20} and obtain the clustering results by tuning the parameters.
The comparison results are shown in Table \ref{result-yale} $\sim$ \ref{result-COIL20}. We compare our approach with the eight methods on each dataset. Moreover, confusion matrices of the t-SVD-MSC and the proposed method is shown in Fig. 1 in \textit{supplemental material}, in which the superiority of the proposed method can be easily observed.
On the six real-world datasets, our approach achieves the best clustering performance in terms of the six criteria. The clustering performance of Kt-SVD-MSC is higher by $(0.0917,0.1052,0.1425,0.1290, 0.1438,0.1140)$ than t-SVD-MSC \cite{t-SVD-MSC} and is higher by $(0.1937,0.2255,0.3295,0.2935,0.3453,0.2243)$ than ECMSC \cite{ECMSC} in terms of the average gain of clustering performance on the six datasets. The results demonstrate that the kernel version of tensor-SVD is helpful for self-representation multi-view clustering, and the proposed method achieves the best performance among nine compared methods.

\begin{figure}[!htb]
\centering
\bf
\subfloat[]{\includegraphics[width=0.25\textwidth]{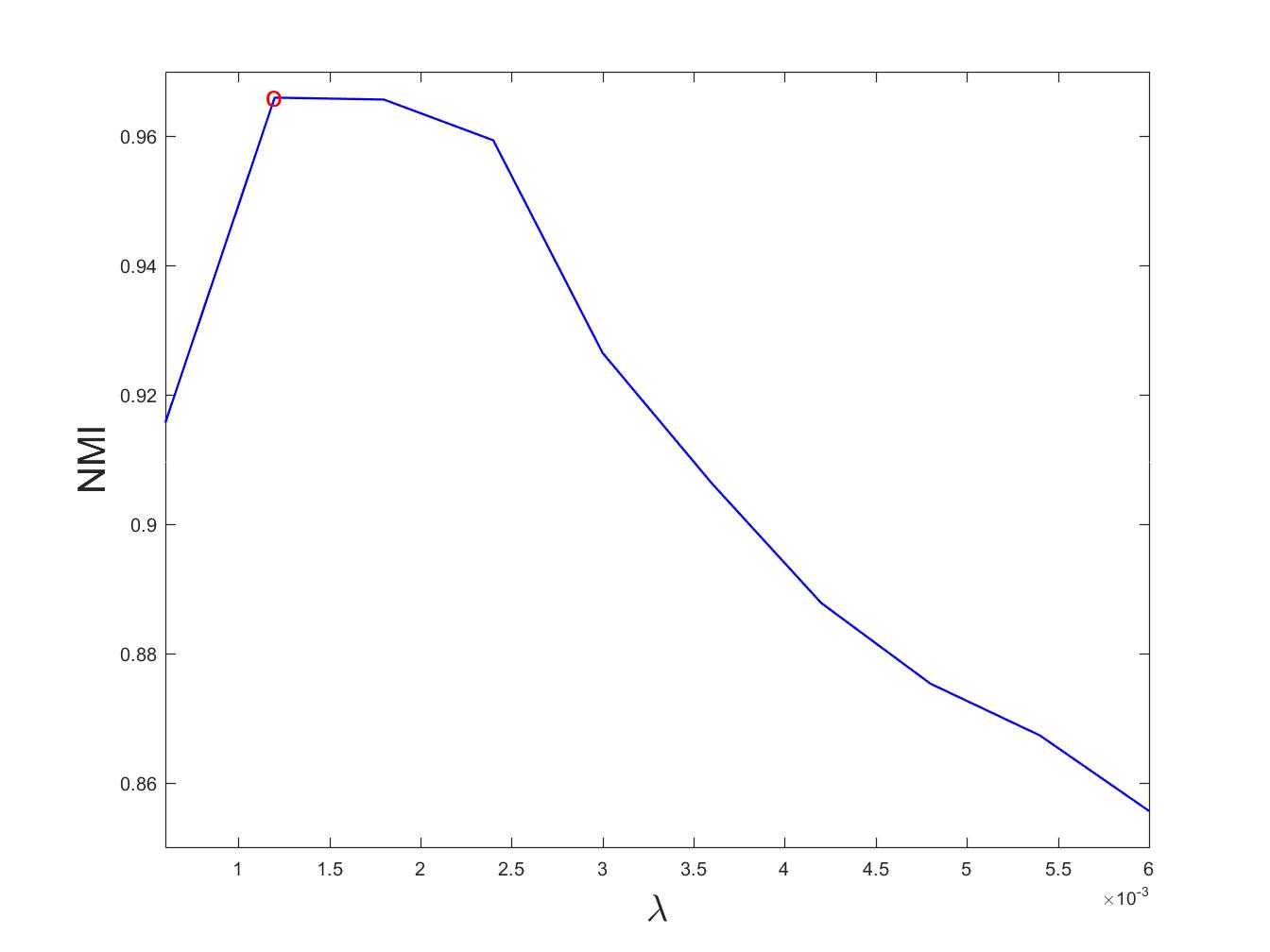}
\label{compare_NMI}}
\subfloat[]{\includegraphics[width=0.25\textwidth]{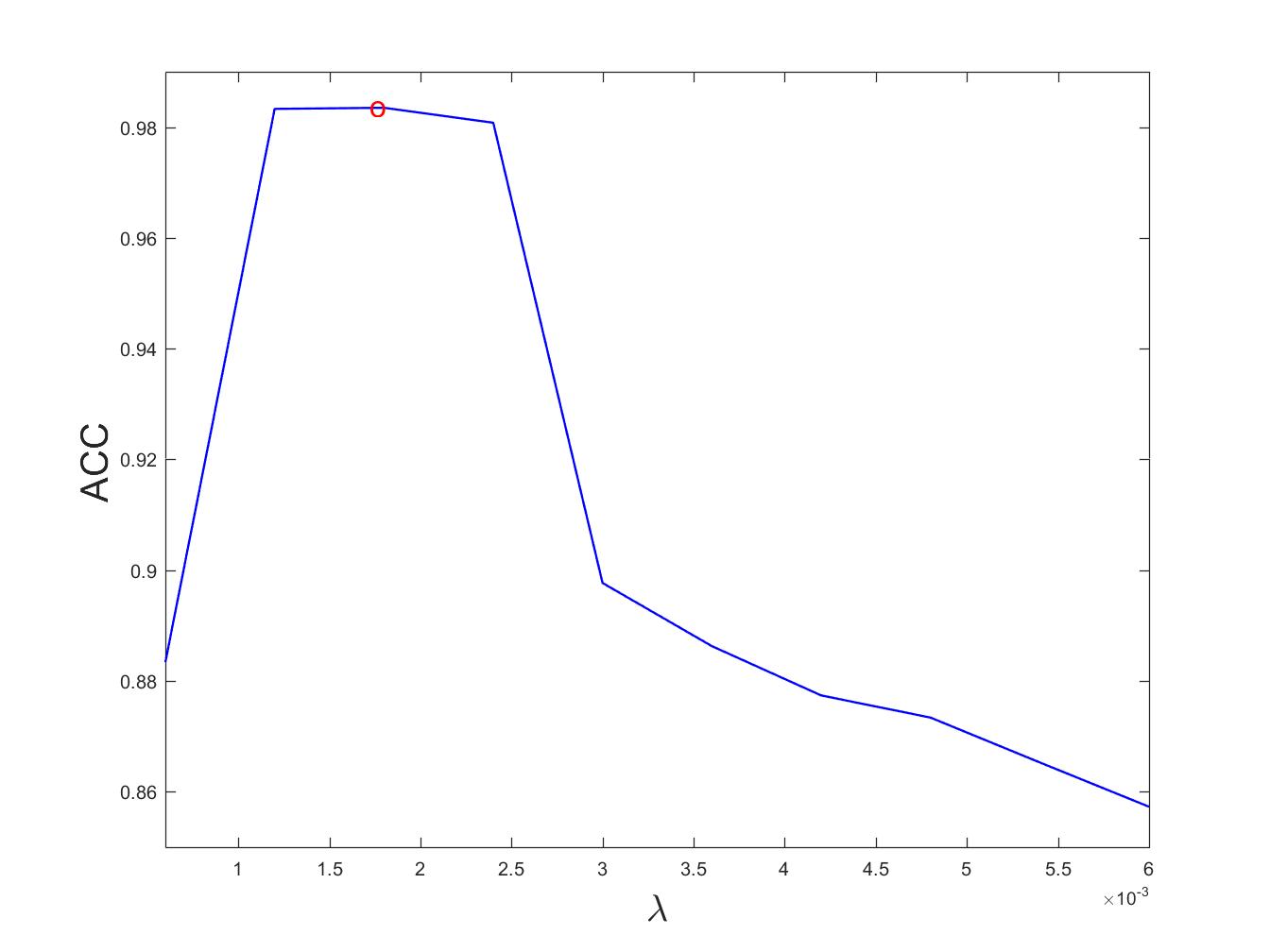}
\label{compare_ACC}}
\caption{The effect of the parameter $\lambda$ to NMI and ACC on {\it Scene-$15$} dataset. a) $\lambda$ affects NMI, b) $\lambda$ affects ACC  }
\label{fig:scene15_NMI_ACC}
\end{figure}

Furthermore, we investigate how the parameter $\lambda$ influences the clustering performance in terms of NMI and ACC. In (\ref{Kt-SVD-MSC-real-problem1}), the proposed model contains only one tuning parameter $\lambda$ which balances the effects of the two parts. And the choice of $\lambda$ is set empirically on the testing data of each dataset. In Fig. \ref{fig:scene15_NMI_ACC}, we plot the curves of NMI and ACC with the change of the parameter $\lambda$. In each curve, a red circle denotes the $\lambda$ value where the clustering performance(NMI or ACC) achieves the highest value. It also demonstrates that most settings of $\lambda$ can make NMI or ACC achieve higher value than the other compared methods. Thus, Fig.\ref{fig:scene15_NMI_ACC} implies that our approach is stable with regards of $\lambda$.

\begin{figure}[htb]
\centering
\includegraphics[width=0.4\textwidth]{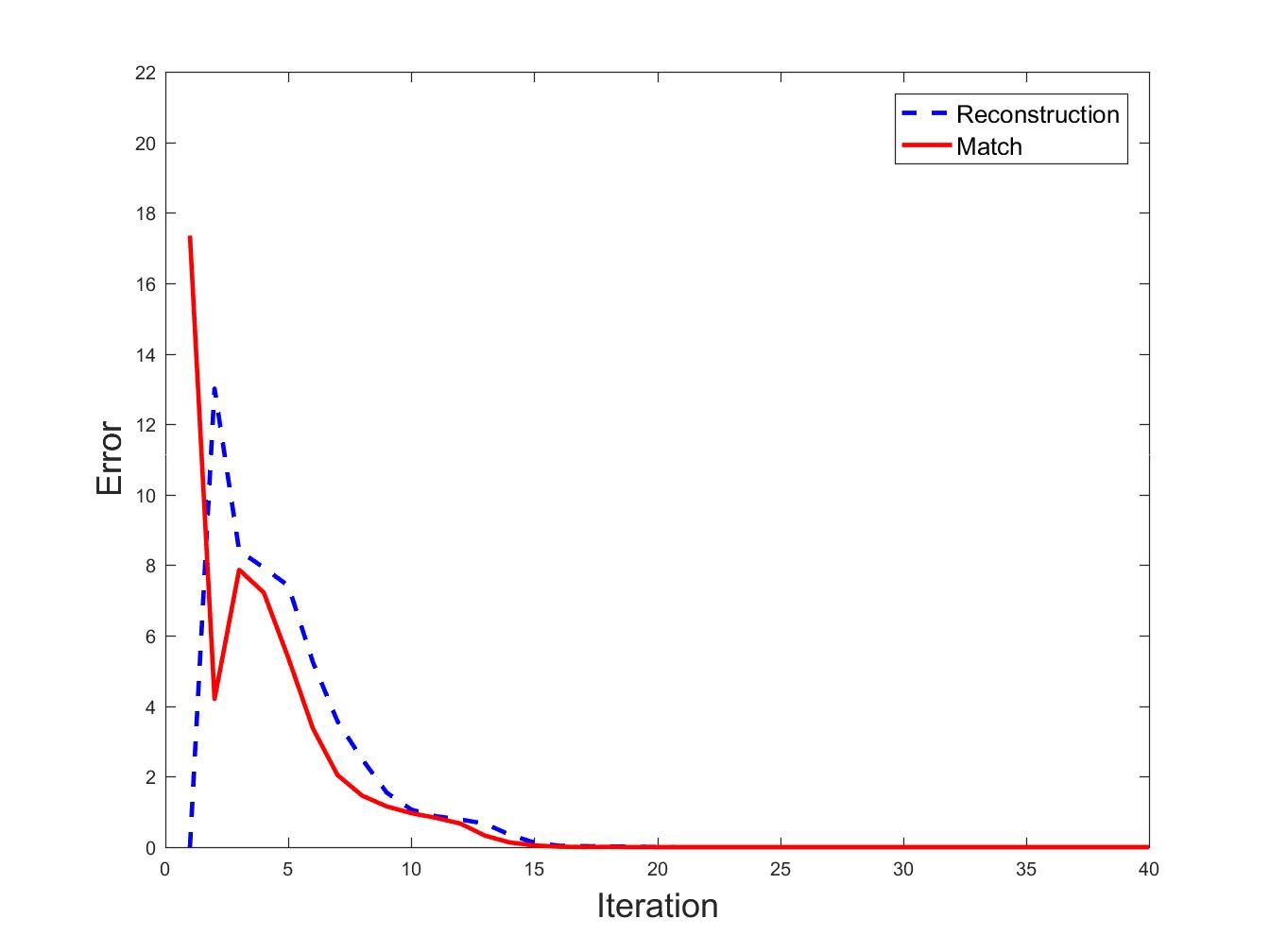}
\caption{Convergence curve on {\it Scene-15}.}
\label{fig:convergence_Scene15}
\end{figure}

Thanks to ADM and the inferred closed-form solution, the proposed optimization model Kt-SVD-MSC converges fast. In order to discuss the convergence of our approach, we define the reconstruction error as (\ref{reconstruction error}) and matching error as (\ref{matching error}). We plot two curves for the two errors on {\it Scene-$15$}. The reconstruction error curve is signed by the solid line and the matching error curve is signed by the dotted line. Observing the changes of the reconstruction error and the matching error with the increase of the iterations in Fig.\ref{fig:convergence_Scene15}, we find that both curves decrease severely to approximate zero during 15 iterations and they become flat after $15$th iteration. Thus, Fig.\ref{fig:convergence_Scene15} demonstrates the proposed Kt-SVD-MSC can converge fast.

\begin{equation}\label{reconstruction error}
 Recon_{error} = \frac{1}{V} \sum_{v=1}^{V}\|X^{(v)}-X^{(v)}Z^{(v)}-E^{(v)}\|_{\infty}
\end{equation}

\begin{equation}\label{matching error}
Match_{error} = \frac{1}{V} \sum_{v=1}^{V}\|Z^{(v)}-G^{(v)}\|_{\infty}
\end{equation}

\begin{table}[!ht]
\footnotesize
\begin{threeparttable}[]
\caption{Clustering results (mean) on {\it Yale}. We set $\lambda = 0.21$ in proposed Kt-SVD-MSC.}\label{result-yale}
\tabcolsep=2pt
\begin{tabular}{@{}l|cccccc}
\toprule[2pt]
Method         &NMI &ACC &AR &F-score &Precision &Recall \\ \hline
$\text{SPC}_{\text{best}}$    &0.654  &0.618  &0.440  &0.475  &0.457  &0.500  \\
$\text{LRR}_{\text{best}}$    &0.709  &0.697  &0.512  &0.547  &0.529  &0.567 \\ \hline
RMSC                          &0.684  &0.642  &0.485  &0.517  &0.500  &0.535 \\
DiMSC                         &0.727  &0.709  &0.535  &0.564  &0.543  &0.586 \\
LTMSC                         &0.765  &0.741  &0.570  &0.598  &0.569  &0.629 \\
Ut-SVD-MSC                    &0.756  &0.733  &0.584  &0.610  &0.591  &0.630 \\
t-SVD-MSC                     &0.953  &0.963  &0.910  &0.915  &0.904  &0.927\\
ECMSC                         &0.773  &0.771 & 0.590  & 0.617 &0.584   &0.653  \\
Kt-SVD-MSC                   &\textbf{0.987}  &\textbf{0.982}  &\textbf{0.973}  &\textbf{0.975}  &\textbf{0.971} &\textbf{0.979} \\
\hline \bottomrule[1pt]
\end{tabular}
\end{threeparttable}
\end{table}

\begin{table}[!ht]
\footnotesize
\begin{threeparttable}[]
\caption{Clustering results (mean) on {\it Extended YaleB}. We set $\lambda = 0.05$ in proposed Kt-SVD-MSC.}\label{result-yaleB}
\tabcolsep=2pt
\begin{tabular}{@{}l|cccccc}
\toprule[2pt]
Method         &NMI &ACC &AR &F-score &Precision &Recall \\ \hline
$\text{SPC}_{\text{best}}$    &0.360   &0.366  &0.225  &0.308  &0.296  &0.310  \\
$\text{LRR}_{\text{best}}$    &0.627   &0.615  &0.451  &0.508  &0.481  &0.539  \\ \hline
RMSC                          &0.157   &0.210  &0.060  &0.155  &0.151  &0.159  \\
DiMSC                         &0.636   &0.615  &0.453  &0.504  &0.481  &0.534  \\
LTMSC                         &0.637   &0.626  &0.459  &0.521  &0.485  &0.539  \\
Ut-SVD-MSC                    &0.479   &0.470  &0.274  &0.350  &0.327  &0.375 \\
t-SVD-MSC                     &0.667   &0.652  &0.500  &0.550  &0.514  &0.590 \\
ECMSC                         & 0.759  &0.783  &0.544  &0.597  &0.513  &0.718 \\
Kt-SVD-MSC                   &\textbf{0.893} &	\textbf{0.896} &\textbf{0.813}  &\textbf{0.832}  &\textbf{0.821} &\textbf{ 0.842 }\\
\hline \bottomrule[1pt]
\end{tabular}
\end{threeparttable}
\end{table}

\begin{table}[!ht]
\footnotesize
\begin{threeparttable}[]
\caption{Clustering results (mean) on {\it ORL}. We set $\lambda = 0.01$ in proposed Kt-SVD-MSC.}\label{result-ORL}
\tabcolsep=2pt
\begin{tabular}{@{}l|cccccc}
\toprule[2pt]
Method         &NMI &ACC &AR &F-score &Precision &Recall \\ \hline
$\text{SPC}_{\text{best}}$    &0.884  &0.725  &0.655  &0.664  &0.610  &0.728  \\
$\text{LRR}_{\text{best}}$    &0.895  &0.773  &0.724  &0.731  &0.701  &0.754  \\ \hline
RMSC                          &0.872  &0.723  &0.645  &0.654  &0.607  &0.709 \\
DiMSC                         &0.940  &0.838  &0.802  &0.807  &0.764  &0.856  \\
LTMSC                         &0.930  &0.795  &0.750  &0.768  &0.766  &0.837  \\
Ut-SVD-MSC                    &0.874  &0.765  &0.666  &0.675  &0.643  &0.708  \\
t-SVD-MSC                     &0.993  &0.970  &0.967  &0.968  &0.946  &0.991\\
ECMSC                         &0.947  &0.854  &0.810  &0.821  &0.783  &0.859 \\
Kt-SVD-MSC                   &\textbf{0.994}  &\textbf{0.971}  &\textbf{0.972}  &\textbf{0.972}  &\textbf{0.956}  &\textbf{0.991} \\
\hline \bottomrule[1pt]
\end{tabular}
\end{threeparttable}
\end{table}

\begin{table}[!ht]
\footnotesize
\begin{threeparttable}[]
\caption{Clustering results (mean) on {\it Notting-Hill}. We set $\lambda = 0.009$ in proposed Kt-SVD-MSC.}\label{result-Notting-Hill}
\tabcolsep=2pt
\begin{tabular}{@{}l|cccccc}
\toprule[2pt] 
Method         &NMI &ACC &AR &F-score &Precision &Recall \\ \hline
$\text{SPC}_{\text{best}}$    &0.723  &0.816  &0.712  &0.775  &0.780  &0.776  \\
$\text{LRR}_{\text{best}}$    &0.579  &0.794  &0.558  &0.653  &0.672  &0.636  \\ \hline
RMSC                          &0.585  &0.807  &0.496  &0.603  &0.621  &0.586  \\
DiMSC                         &0.799  &0.837  &0.787  &0.834  &0.822  &0.827  \\
LTMSC                         &0.779  &0.868  &0.777  &0.825  &0.830  &0.814  \\
Ut-SVD-MSC                    &0.837  &0.933  &0.847  &0.880  &0.900  &0.861  \\
t-SVD-MSC                     &0.900  &0.957  &0.900  &0.922  &0.937  &0.907  \\
ECMSC                         &0.817  &0.767  &0.679  &0.764   &0.637  & 0.954  \\
Kt-SVD-MSC                   &\textbf{0.973}&\textbf{0.992}	&\textbf{0.981}	&\textbf{0.985}	&\textbf{0.989}	&\textbf{0.981}\\
\hline \bottomrule[1pt]
\end{tabular}
\end{threeparttable}
\end{table}

\begin{table}[!ht]
\footnotesize
\begin{threeparttable}[]
\caption{Clustering results (mean) on {\it Scene-15}. We set $\lambda = 0.0018$ in proposed Kt-SVD-MSC.}\label{result-Scene15}
\tabcolsep=2pt
\begin{tabular}{@{}l|cccccc}
\toprule[2pt] 
Method         &NMI &ACC &AR &F-score &Precision &Recall \\ \hline
$\text{SPC}_{\text{best}}$    &0.421  &0.437  &0.270  &0.321  &0.314  &0.329 \\
$\text{LRR}_{\text{best}}$    &0.426  &0.445  &0.272  &0.324  &0.316  &0.333 \\ \hline
RMSC                          &0.564  &0.507  &0.394  &0.437  &0.425  &0.450 \\
DiMSC                         &0.269  &0.300  &0.117  &0.181  &0.173  &0.190 \\
LTMSC                         &0.571  &0.574  &0.424  &0.465  &0.452  &0.479 \\
Ut-SVD-MSC                    &0.555  &0.510  &0.375  &0.422  &0.389  &0.460 \\
t-SVD-MSC                     &0.858  &0.812  &0.771  &0.788  &0.743  &0.839 \\
ECMSC                         &0.463  &0.457  &0.303  &0.357  &0.318  &0.408  \\
Kt-SVD-MSC                   &\textbf{0.966} &\textbf{0.984}  &\textbf{0.967} &\textbf{0.969}&\textbf{0.971}&\textbf{0.968} \\
\hline \bottomrule[1pt]
\end{tabular}
\end{threeparttable}
\end{table}

\begin{table}[!ht]
\footnotesize
\begin{threeparttable}[]
\caption{Clustering results (mean) on COIL-20. We set $\lambda = 0.0009$ in proposed Kt-SVD-MSC.}\label{result-COIL20}
\tabcolsep=2pt
\begin{tabular}{@{}l|cccccc}
\toprule[2pt] 
Method         &NMI &ACC &AR &F-score &Precision &Recall \\ \hline
$\text{SPC}_{\text{best}}$    &0.806  &0.672  &0.619  &0.640  &0.596  &0.692 \\
$\text{LRR}_{\text{best}}$    &0.829  &0.761  &0.720  &0.734  &0.717  &0.751 \\ \hline
RMSC                          &0.800  &0.685  &0.637  &0.656  &0.620  &0.698 \\
DiMSC                         &0.846  &0.778  &0.732  &0.745  &0.739  &0.751 \\
LTMSC                         &0.860  &0.804  &0.748  &0.760  &0.741  &0.776 \\
Ut-SVD-MSC                    &0.841  &0.788  &0.732  &0.746  &0.731  &0.760 \\
t-SVD-MSC                     &0.884  &0.830  &0.786  &0.800  &0.785  &0.808 \\
EMSC                                      & 0.942  & 0.782 &0.781  &0.794  &0.695  &0.925  \\
Kt-SVD-MSC                                &\textbf{0.992} &\textbf{0.990}&\textbf{0.983} & \textbf{0.984} & \textbf	{0.984}&\textbf{0.985} \\
\hline \bottomrule[1pt]
\end{tabular}
\end{threeparttable}
\end{table}

\section{Conclusions}\label{sec:conclusion}

In this paper, we proposed a kernel version of self-representation multi-view clustering with tensor multi-rank minimization. A new optimization model for the robust kernel tensor-SVD representation is made. After that, ADM algorithm is implemented on solving this optimization problem. We provide the closed-form solutions for the $l_{2,1}$ norm-related subproblems so that the subproblems can be efficiently solved. Extensive experimental results on three types of applications: face clustering, scene clustering, and generic object clustering have clearly demonstrated that the proposed self-representation multi-view clustering method based on kernelized tensor-SVD is effective and efficient. In the nine compared clustering methods, our approach achieves the breakthrough clustering performance on Scene-15 which is a challenging clustering dataset and achieves the state-of-the-art clustering results on the other datasets.
%

\end{document}